\documentclass[lettersize,journal]{IEEEtran}
\usepackage{amsmath,amsfonts}
\usepackage{algorithmic}
\usepackage{multirow}
\usepackage{algorithm}
\usepackage{array}
\usepackage[caption=false,font=normalsize,labelfont=sf,textfont=sf]{subfig}
\usepackage{textcomp}
\usepackage{stfloats}
\usepackage{url}
\usepackage{verbatim}
\usepackage{graphicx}
\def\BibTeX{{\rm B\kern-.05em{\sc i\kern-.025em b}\kern-.08em
    T\kern-.1667em\lower.7ex\hbox{E}\kern-.125emX}}
\usepackage{balance}
\usepackage{cite}
\begin{document}
\title{Enhanced Prediction of Multi-Agent Trajectories via  Control Inference and State-Space Dynamics}
\author{Yu Zhang, Yongxiang Zou, Haoyu Zhang, Zeyu Liu, Houcheng Li, Long Cheng~\IEEEmembership{Fellow,~IEEE,}
\thanks{This work was supported in part by the National Natural Science Foundation of China (Grants 62025307, U1913209) and was also supported by the CAS Project for Young Scientists in Basic Research (Grant No. YSBR-034).}\thanks{The authors are all with the School of Artificial Intelligence, University of Chinese Academy of Sciences, Beijing 100049, China. They are also with the State Key laboratory of Multimodal Artificial Intelligence Systems, Institute of Automation, Chinese Academy of Sciences, Beijing 100190, China. All correspondences should be addressed to the corresponding author Dr. Long Cheng (long.cheng@ia.ac.cn).}}

\markboth{Journal of \LaTeX\ Class Files,~Vol.~18, No.~9, September~2020}%
{How to Use the IEEEtran \LaTeX \ Templates}

\maketitle

\begin{abstract}
In the field of autonomous systems, accurately predicting the trajectories of nearby vehicles and pedestrians is crucial for ensuring both safety and operational efficiency. This paper introduces a novel methodology for trajectory forecasting based on state-space dynamic system modeling, which endows agents with models that have tangible physical implications. To enhance the precision of state estimations within the dynamic system, the paper also presents a novel modeling technique for control variables. This technique utilizes a newly introduced model, termed "Mixed Mamba," to derive initial control states, thereby improving the predictive accuracy of these variables. Moverover, the proposed approach ingeniously integrates graph neural networks with state-space models, effectively capturing the complexities of multi-agent interactions. This combination provides a robust and scalable framework for forecasting multi-agent trajectories across a range of scenarios. Comprehensive evaluations demonstrate that this model outperforms several established benchmarks across various metrics and datasets, highlighting its significant potential to advance trajectory forecasting in autonomous systems.
\end{abstract}


\section{Introduction}
\IEEEPARstart{T}{he} transition of autonomous vehicles (AVs) from a futuristic vision to a practical reality is accelerating, with technological advancements enabling trials and limited public road deployments \cite{b1}. However, the progression towards fully operational self-driving cars hinges on a natural comprehension of social norms and behaviors. The effectiveness of this understanding hinges on the capability to reliably predict the behaviors of other road users.
The complex interplay among traffic participants, wherein the actions of an individual can markedly influence the decision-making of others, highlights the imperative to refine predictive models. These models are essential for the robust development of autonomous driving strategies \cite{b2,b3,b4,b5,bu1,bu2}.
\begin{figure}[htbp]
	\centerline{\includegraphics[width=0.5\textwidth]{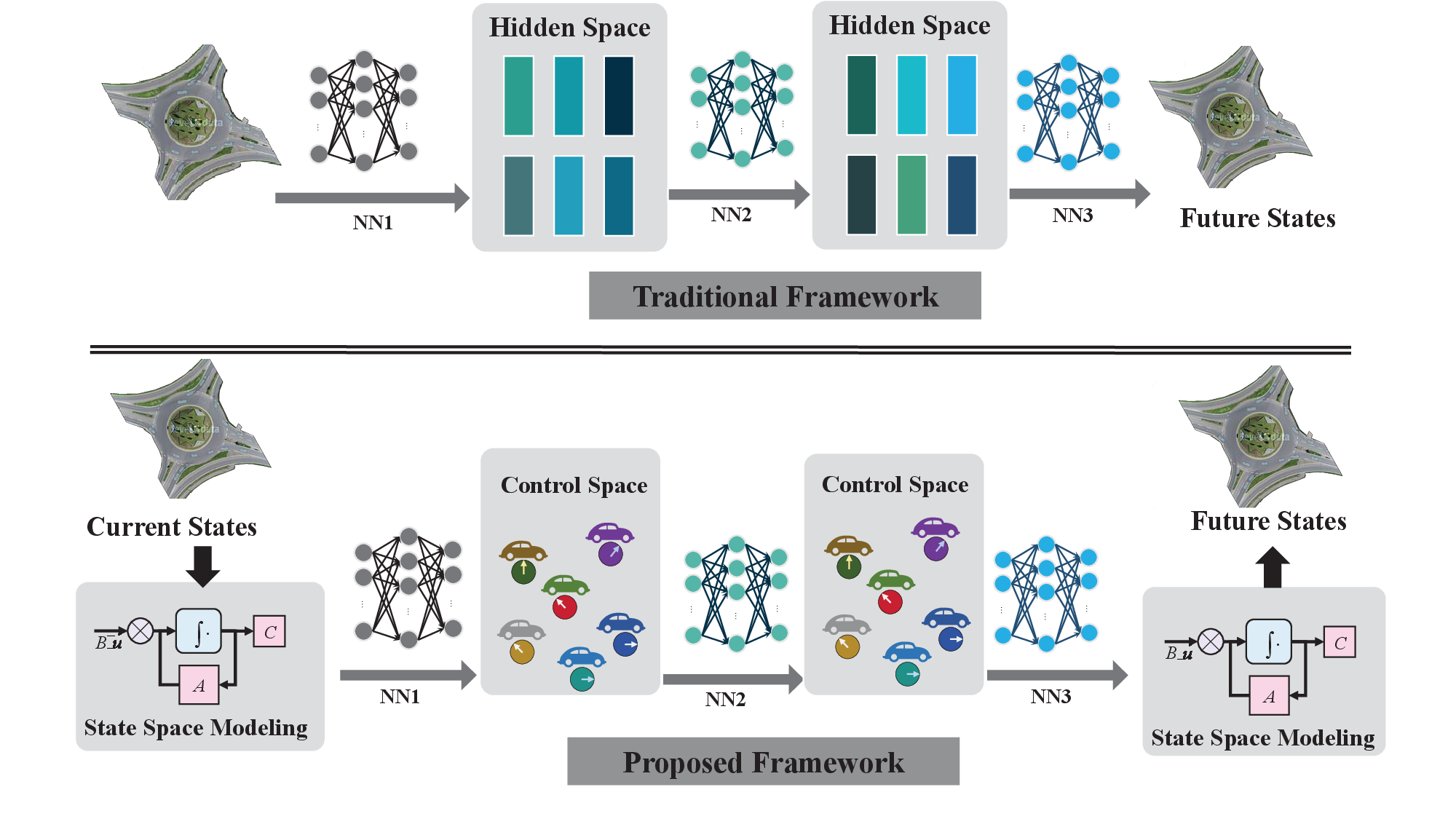}}
	\caption{The main difference between the proposed framework and traditional frameworks for multi-agent trajectory prediction.}	
	\label{figbu}
\end{figure}
As sensing technology progresses and computing power escalates, there is a corresponding surge in the availability of high-quality data. The convergence of these advancements significantly amplifies the capacity to adopt data-driven methodologies for behavior prediction. Notably, the vision-based systems integral to many vehicles have already endowed them with considerable computational capabilities, thereby facilitating the implementation of such predictive analytics.  The utilization of Graph Neural Networks (GNNs) has risen as a particularly effective strategy for addressing the problem under consideration. This is largely attributable to their robust relational inductive bias. Additionally, GNNs inherently excel at forecasting the behavior of multiple agents due to their ability to represent and reason about the relationships among participants within the problem domain. \cite{b6}.
Moreover, the ability of GNNs to forecast the behavior of multiple agents is a natural outcome of their structure. By representing road participants as nodes within a graph, GNNs can generate simultaneous predictions for numerous targets. This approach significantly enhances the comprehensiveness of forecasting model.

Although Transformers have significantly improved modeling capabilities, their application to sequence data is often impeded by the quadratic computational cost inherent in their attention mechanism. In contrast, RNN-based models are inherently adept at processing sequential information. To address the issue of context compression in sequence modeling, Mamba, a selective State Space Model (SSM), has been introduced \cite{Mamba}. Unlike the attention-based approach, Mamba leverages the framework of SSMs to encode context through hidden states during recurrent iterations. A pivotal feature of Mamba is its selection mechanism, which facilitates the precise control over input segments that contribute to the formation of hidden states, thereby exerting a consequential influence on subsequent embedding updates. This selection process can be conceptualized as a data-driven node selection in graph modeling, where Mamba selectively filters relevant nodes at each recurrent step, focusing only on the chosen context. This targeted approach to context sparsification aligns with the objectives of attention sparsification and presents a viable alternative to random subsampling techniques. Furthermore, the Mamba module is meticulously engineered to achieve linear time complexity and a minimized memory footprint, thereby significantly enhancing efficiency in large-scale graph training tasks.

Despite their versatility and superior performance across a range of tasks, deep neural networks often fail to match the interpretability offered by conventional state-estimation methods. However, recent advancements have shown promise by applying deep neural networks to determine the optimal inputs for an intrinsic motion model, which in turn improves interpretability by ensuring that predictions are consistent with physical laws \cite{b7,b8,b9}. These strategies are well-founded, drawing from an extensive body of literature on motion modeling \cite{b10,b11}.
While the dynamics of different road users, including pedestrians, bicycles, and cars, vary considerably, this diversity may necessitate a distinct set of models for each. A promising solution to this challenge is the application of a more adaptable class of methods. One such approach is the use of Neural Ordinary Differential Equations (neural ODEs) \cite{b12}. Neural ODEs provide the adaptability to model and learn the intrinsic differential constraints unique to each category of road user, as utilized in the original MTP-GO \cite{b3}. However, directly modeling may result in poor generalization performance. In contrast, this paper use the SSM model to addresses this challenge. 

As depicted in Figure \ref{figbu}, the proposed framework diverges significantly from the conventional data-driven techniques, which rely on a two-step neural network process. These techniques map agent trajectories into an implicit latent space and then evolve these states to project future positions.

In contrast, the approach initiates with SSM to encapsulate the dynamics of multiple agents.  When appropriately designed, SSM modeling can be viewed as a specific case of Koopman modeling.  , offering a structured perspective on system dynamics.

Employing a neural network to predict the control variables that is essential for the SSM. These variables are then further evolved within the control space by another neural network. The process concludes with the computation of the predicted agent states through the SSM, harnessing the fully evolved control variables.

A key advantage of the proposed framework is the tangible physical meaning conveyed by the control variables. This is in marked contrast to the elusive implicit states characteristic of traditional methods, which often lack clear physical interpretations. The main contribution of the paper are listed as follows:
\begin{itemize}
	\item SSM dynamic system: In this study,  a DS approach to model the behavior of agents is used, offering a nuanced and flexible framework for the analysis of complex and nonlinear phenomena. The proposed methodology integrates the use of SSM, which effectively harnesses the precision of linear analysis techniques while addressing the intrinsic nonlinearity of the systems under investigation. By doing so,  both the predictive accuracy and the interpretability of the model are enhanced, thereby providing deeper insights into the underlying dynamics
	
	\item Selection state space: GNNs are adept at capturing the spatial relationships among various agents.  While the selection mechanism are proficient in capturing temporal sequences, combining Mamba with the graph neural networks help to catch the spatio-temporal relationship. A unique mixed Mamba structure is proposed to augment the system's analytical capabilities.
	
	\item State space modeling of control variables: In this paper, the control variables are formulated as SSM contingent upon the estimated past states and predicted states of the agents. These evolving control variables are then integrated into the DS model, which in turn influences the evolution of the agents' states.
		
	\item The proposed algorithm demonstrates superior performance over several state-of-the-art methods  across three distinct public datasets.
\end{itemize}

The structure of this paper is outlined as follows: Section \ref{sec2} provides an exhaustive examination of the pertinent literature. Section \ref{sec3} delves into the detail of the proposed framework to predict multi-agent trajectories. In Section \ref{sec4},  a comparative analysis of the performance of the refined algorithm against several state-of-the-art methods  is presented, showcasing the results of the evaluations across a spectrum of metrics and datasets. Finally, Section \ref{sec5} encapsulates the findings and concludes the paper, offering insights into the implications of the research and potential directions for studies.

\section{RELATED WORK} \label{sec2}
\subsection{Traffic Behavior Prediction} 
Over the past decade, the field of behavior prediction has garnered substantial attention within the academic community. Thorough insights and extensive analyses on this topic are meticulously presented in several research surveys \cite{b14,b15,b16}. Broadly speaking, the field of behavior prediction is typically divided into three primary categories: intention prediction, motion prediction, and social interactions prediction. These categories often utilize sequential input data, such as historical agent positions, to forecast outcomes. Recurrent Neural Networks (RNNs), and notably  Long Short-Term Memory (LSTM) networks, are commonly employed in the domain of this predictive task.

The principal aim of intention prediction methodologies is to infer high-level decision-making processes that are intrinsic to the dynamics of the traffic scene. This encompasses the anticipation of actions at intersections, as discussed in \cite{b17}, as well as the estimation of the probability of lane changes on highways, as detailed in \cite{b18}. Within these methodologies, agent trajectories are meticulously annotated in accordance with user-defined maneuvers, and predictive models are subsequently constructed within a supervised learning paradigm.

While intention prediction is fundamentally characterized as a classification problem, in contrast, motion prediction is inherently framed as a regression task, with distance-based metrics commonly serving as the learning objectives. However, these two predictive endeavors are not mutually exclusive. This interconnection is exemplified in the work of \cite{b19}, where the predicted intention informs the anticipation of future trajectories. Given the sequential essence of the motion prediction challenge, numerous methods have adopted the encoder-decoder architecture as the foundation for their models \cite{b20,b21}.

Historically, the domain of social interaction modeling has predominantly relied upon model-driven approaches, which leverage intricate mathematical and statistical models to delineate and understand the motion patterns exhibited by agents within a given system. These structured frameworks have been instrumental in analyzing and predicting agent behavior across various contexts \cite{b22,b23}. In recent years, as deep learning techniques have progressed continuously, data-driven artificial intelligence  methodologies have gained prominence within the domain of trajectory prediction \cite{b15,b24,b25}. 
Notably, the Sequential Generative Adversarial Network (SGAN) \cite{b26} employed Generative Adversarial Networks (GAN) to forecast multimodal trajectories and to introduce an innovative pooling mechanism that calculates social interactions based on the relative spatial relationships among pedestrians. 
Additionally, Graph Convolutional Networks (GCNs) have emerged as a prevalent approach for accurately forecasting the trajectories of agents across diverse scenarios \cite{gcn}. For instance, the Social Spatial-Temporal Graph Convolutional Network (SSTGCNN)  \cite{b27}
incorporates social interactions among pedestrians into an adjacency matrix, thereby treating them within the framework of a graph network structure. This approach leverages kernel functions to process the adjacency matrix, effectively capturing spatial and temporal interaction information. \cite{b28} introduced a mixed graph convolutional network designed to holistically account for the aforementioned factors, with the objective of enhancing the accuracy of multi-class trajectory prediction. This approach involves the construction of a multi-type graph-guided adjacency matrix that encapsulates the various interaction relationships present within the scene.

\subsection{GNNs for Trajectory Prediction} 
GNNs constitute the deep learning models that are specifically tailored for processing graph-structured data \cite{b29}. When provided with a graph structure and an associated set of features, GNNs are capable of learning meaningful representations for individual nodes, edges, or the entire graph. \cite{b30}. These learned features serve as a foundation for a myriad of prediction tasks, enabling GNNs to be applied across a diverse array of applications 
Temporal GNNs are designed to analyze graph-structured data that is collected sequentially over time. This data forms time series, where each point in the series is associated with a graph. Therefore, temporal GNNs extend the capabilities of traditional GNNs by incorporating mechanisms that explicitly account for the temporal evolution within the data. These models are capable of integrating various models to capture temporal patterns effectively. Include RNNs \cite{b31}, Convolutional Neural Networks (CNNs) \cite{b32}, or attention mechanisms \cite{b7}. Traditionally, the majority of temporal GNNs function on the premise of fixed and known graph structures, as discussed in \cite{b32,b33}. Meanwhile, recent research has ventured into the dynamic learning of graph structures themselves, as explored in \cite{b34}.
Furthermore, GNNs are particularly adept at addressing trajectory prediction problems. This is achieved by conceptualizing the edges within the graph as representations of interactions among agents. The proposed model in \cite{b35}, employs an encoder-decoder framework to forecast the trajectories of physical objects that are in interaction with each other. Within this framework, a GNN is utilized to encode historical information of the system's state, and subsequently a neural
ODE decoder is applied to forecast the future trajectories. In the study  \cite{b36}, the application of diverse GNNs to model interactions among traffic participants for motion prediction was explored. Despite being an early foray into this area, the research yielded encouraging outcomes for Interaction-aware (IA) modeling.
GRIP++, as introduced in \cite{b37}, is a graph-structured recurrent framework specifically tailored for the prediction of vehicle trajectories. It translates the scene into a latent representation by employing layers of GNNs, as discussed in \cite{b38}, the latent representation is subsequently fed into a RNN-based framework for the purpose of trajectory forecasting.
The SCALE-net, as discussed in \cite{b39}, demonstrates proficiency in managing arbitrary interacting agents. Unlike GRIP++, SCALE-net employs an attention mechanism, which is influenced by graph edge features, to encode node feature updates, as elaborated in \cite{b40}.
In \cite{b41}, a method was proposed for learning node-wise interactions using solely a graph-attention mechanism, as described in \cite{b42}. Following this, the graph encoding is channeled into a LSTM-based decoder, which is utilized for the prediction of vehicle trajectories. 
The MTP-GO model \cite{b3}, however, is motivated by the significance of interaction-aware features and consistently maintains the graph structure throughout the entire prediction process. This approach ensures that crucial interactions are preserved for the entire duration of the prediction.
\subsection{Applications of State Space Models
}
SSMs serve as a fundamental scientific framework across various disciplines, including control theory and computational neuroscience. These models are characterized by a straightforward equation that maps control signals $\boldsymbol{u}_{t}$ to an underlying state $\boldsymbol{x}_{t}$, which is then projected to generate output signals $\textbf{}$. The latent states within SSMs are often represented using Hidden Markov Models (HMMs), providing a robust mechanism for modeling and analyzing dynamic systems.
	\begin{equation}\label{equ1}
		\begin{cases}
\boldsymbol{x}'_{t}=\boldsymbol{A}\boldsymbol{x}_{t}+\boldsymbol{B}\boldsymbol{u}_{t} \\
\boldsymbol{y}_{t}=\boldsymbol{C}\boldsymbol{x}_{t}
		\end{cases}
\end{equation}
When the parameters $\boldsymbol{A}, \boldsymbol{B}$, and $\boldsymbol{C}$ in (\ref{equ1}) are also time-variant, the SSM evolves into Mamba.
Mamba introduces an innovative approach that efficiently captures long-range dependencies with a linear computational complexity. This method effectively addresses the computational inefficiencies typically associated with Transformers, while fully preserving their ability to model global information. Due to its excellent performance, Mamba has rapidly garnered significant attention from a substantial number of researchers in the field. U-Mamba \cite{b43} and VM-UNet \cite{b44} have integrated the Mamba block into the U-net architecture, specifically tailoring it for addressing challenges in biomedical image segmentation. To augment the flow of information on a per-channel basis, the MambaMixer architecture, as detailed in \cite{b45}, has been further developed to include channel-mixing Mamba blocks. This enhancement has successfully broadened the application scope of MambaMixer to encompass not only image recognition but also time series forecasting.
 However, the current U-shaped Mamba architectures lack the integration of channel-wise Self-Similarity Module (SSM) components. These modules are crucial for effectively compressing and reconstructing features, especially when dealing with the rich contextual information present within channel dimensions.
 \cite{b46} introduce an innovative dual-directional Mamba U-Net model that addresses this gap. Our proposed model simultaneously considers both global contextual information and channel correlations, offering an efficient and effective approach to image restoration.

\section{Preliminaries} \label{sec3}

In the realm of model-based control theory, the precise representation of the dynamics between the system's input
$u$, and its state, represented by 
$x$, is fundamental for constructing robust and efficient control models. These models are often differentiated into two types: the forward model and the inverse model. The forward model, denoted as $f$, projects the current state  $x_t$ and control input $u_t$ onto the subsequent state $x_{t+1}$. Conversely, the inverse model, denoted as $f^{-1}$,  infers the required control input $u_t$ from the current state $x_t$ and system next state $x_{t+1}$. This relationship can be mathematically formalized as follows:
\begin{equation}
f(\boldsymbol{x}_{t},\boldsymbol{u}_{t};\theta) = \boldsymbol{x}_{t+1}, f^{-1}(\boldsymbol{x}_{t},\boldsymbol{x}_{t+1};\theta) = \boldsymbol{u}_{t},	
\end{equation}
where $\theta$ represents the parameters of the model.

Here's a refined version of your academic writing:

Upon completing the model construction, it is crucial to apply an appropriate numerical integration technique. Several options are available: the explicit Euler method, which is known for its simplicity; the Runge–Kutta methods, which strike a balance between accuracy and computational efficiency; and symplectic integrators, which are especially advantageous for preserving the phase space structure in Hamiltonian systems. These integrators enable accurate prediction of the system's future state, rather than just estimating state transitions.
\subsection{Model engineering} 

The primary methodology in robotics is model engineering, which is widely used across various industries. This approach involves using the transfer function  
$f$, which represents the equations of motion, along with model parameters that describe the robot's physical characteristics, such as mass, center of gravity, dimensions, and moments of inertia. Deriving these equations of motion is a manual process tailored to each robotic system. It typically requires assuming that the robot consists of ideal rigid bodies connected by perfect joints and applying Newtonian, Lagrangian, or Hamiltonian mechanics, in conjunction with the system's structure, to formulate the equations.

Additionally, the parameters for both the forward and inverse models are consistent, allowing for the simultaneous development of both models. This methodology can produce highly accurate forward and inverse models for systems composed of rigid bodies. However, it demands considerable effort, as the parameters must be carefully determined.
\subsection{ Data-driven system identification}

Similar to model engineering, data-driven system identification also uses the analytic equations of motion as the transfer function. However, in contrast to model engineering, the model parameters in data-driven system identification are derived from observed data rather than direct measurements. This approach still requires the manual derivation of the equations of motion, but the parameters are learned through data analysis rather than being directly measured. 

Simple dynamic parameters can be extracted using linear regression with hand-designed features. This method is commonly used as the standard system identification technique for robot manipulators. However, it has limitations. It does not inherently guarantee that the derived parameters are physically plausible, as dynamic parameters must satisfy specific constraints. For example, this approach may result in unrealistic outcomes such as negative masses, an inertia matrix that is not positive definite, or violations of the parallel axis theorem.

The drawbacks of this approach include its inability to infer more than linear combinations of dynamic parameters. Formulating inverse dynamics can also be problematic, as it may not have a unique solution, particularly due to friction \cite{Ratliff}.

To address these issues, some researchers have proposed projection-based approaches, while others have employed virtual parameterizations to ensure physically plausible parameters \cite{Sutanto, Lutter}. The latter approach involves a more complex optimization process than linear regression, often requiring gradient descent for solution.

In summary, this data-driven approach relies solely on the analytic form of the equations of motion and learns the dynamical parameters from observational data. Although it is less labor-intensive than model engineering, collecting high-quality data is crucial for effective learning.
\subsection{ Black-box model learning} 
In contrast to traditional methods that require a detailed understanding of the individual kinematic chain to derive the equations of motion, black-box approaches operate without prior knowledge of the system. These methodologies use any black-box function approximator as the transfer function and adjust the model parameters to fit the observed data.

For instance, the existing literature has utilized various techniques such as  Gaussian Mixture Models (\cite{gmm1},\cite{gmm2}), Gaussian Processes (\cite{gp1},\cite{gp2}),  feedforward networks (\cite{zy1},\cite{zy2}), and recurrent neural networks (\cite{Hafner}) to learn the dynamics model.

Black-box models typically focus on learning either the forward or inverse model, and their validity is confined to the distribution of the training data. In contrast, traditional methods based on analytic equations of motion can derive both models simultaneously and generalize beyond the data distribution, as the learned physical parameters are globally applicable.

However, black-box models have the advantage of not requiring prior assumptions about the system, enabling them to learn systems that involve contacts. Traditional approaches, which rely on rigid body dynamics, can only model articulated bodies using reduced coordinates and without considering contacts. Consequently, black-box models can potentially offer more accurate predictions for real-world systems where the underlying assumptions of traditional methods are not valid. Despite this, black-box models are generally limited to the training domain and rarely extrapolate beyond it.
\section{The Proposed Model} \label{sec4}
\begin{figure*}[htbp]
	\centerline{\includegraphics[width=1.15\textwidth]{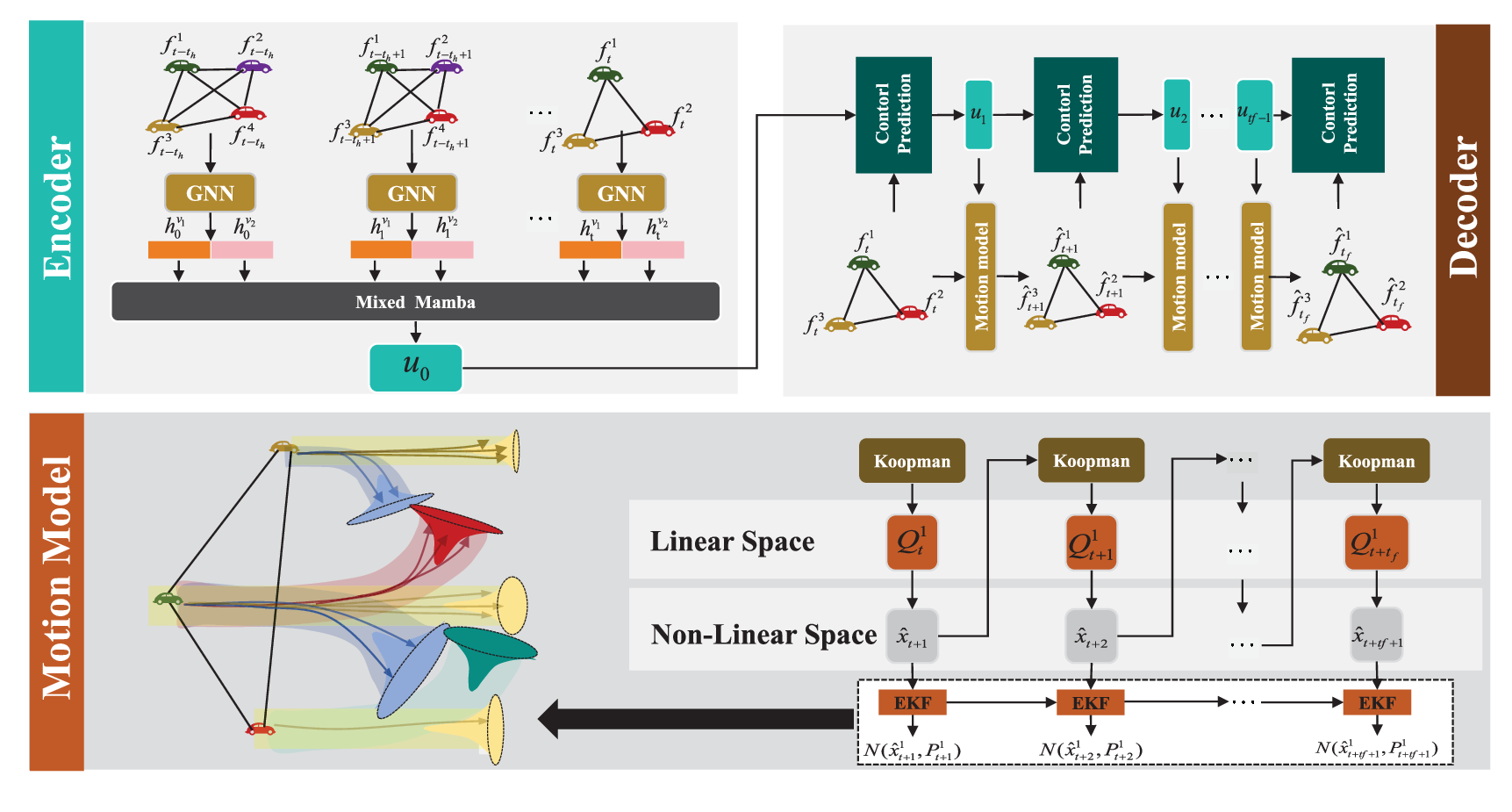}}
	\caption{The overall structure of the proposed algorithm.}	
	\label{fig1}
\end{figure*}

\begin{figure}[htbp]
	\centerline{\includegraphics[width=0.5\textwidth]{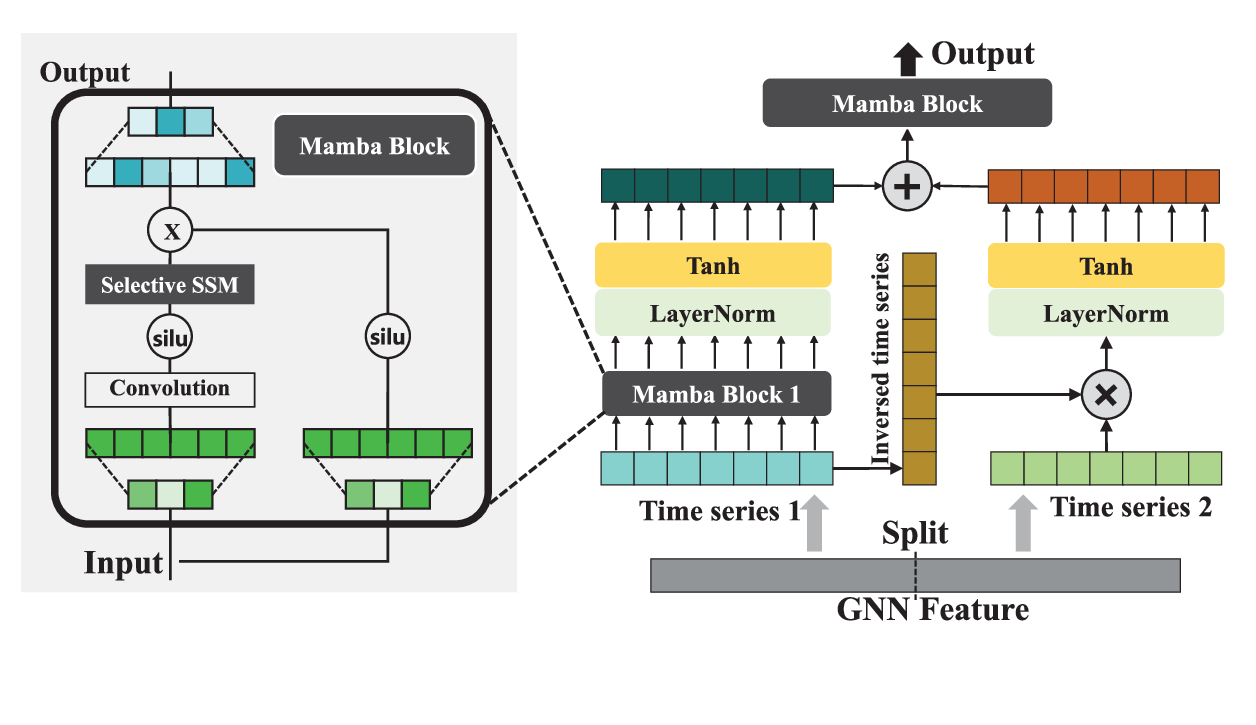}}
	\caption{The overall structure of the proposed mixed Mamba.}	
	\label{fig2}
\end{figure}

Nowadays, the multi-agent trajectory prediction models \cite{b3,b37,b39,b41} integrate the GNN-based encoder-decoder module, which processes historical observations and feeds them into a neural motion model to facilitate trajectory forecasting have been proven effective.  In this paper, this framework is continually utilized but is introduced from different views. 
\subsection{Dynamic modeling} \label{sec4_1}

This is originated from that multi-agent systems can be viewed as a collection of individual systems, each operating under the influence of physical laws. These systems' dynamics can be precisely captured through the use of differential equations. The specific differential equation that models the behavior of such a system is presented as follows:

\begin{equation}\label{equ3}
	\boldsymbol{\dot{x}} = \boldsymbol{A}\boldsymbol{x}+\boldsymbol{B} \boldsymbol{u},	
\end{equation}
where  $\boldsymbol{A}\boldsymbol{x}$ can be seen as the free response for the system while $\boldsymbol{B} \boldsymbol{u}$ denotes the response due to external forces.
Due to the inherent physical characteristics of the system's dynamics, both $\boldsymbol{A}$ and $\boldsymbol{B}$
 remain constant over time. These properties can be effectively modeled using neural networks and learned through data-driven approaches. Which can be enforced by the following loss function:
$||\dot{\boldsymbol{x}_{t}}-	\boldsymbol{A}\boldsymbol{x}_{t}-\boldsymbol{B}\boldsymbol{u}_{t}||_2$, where $\boldsymbol{A}$ and $\boldsymbol{B}$ are a matrices learned automatically, $||\cdot||_2$ denotes the $L_2$-norms.
 
In comparison to direct using SSM model,  
the Koopman operator has emerged as a pivotal mathematical instrument within an array of scientific fields. This operator transforms the nonlinear dynamics into a linear representation by defining a state space through observable functions. The adoption of the Koopman operator framework not only endows the model with interpretability but also streamlines the automatic derivation of observable functions. This is accomplished through the integration of deep neural networks, a strategy corroborated by recent scholarly work \cite{koopman}.

In the realm of unforced dynamics, \cite{b47} introduces an autoencoder framework designed to learn a dictionary of observable functions. This is achieved by approximating the Koopman operator, a key concept in dynamical systems theory, using a finite-dimensional representation. The approximation process leverages a least-squares optimization approach to ensure accuracy and efficiency in the derivation. The Koopman operator, an infinite-dimensional linear transformation, provides a powerful framework for predicting the future trajectories of nonlinear dynamical systems, which acts on a space of observable functions. By decomposing the Koopman operator spectrally, one can uncover the intrinsic dynamics of the system. This decomposition reveals the eigenvalues and eigenfunctions that correspond to the system's behavior, offering a deeper understanding of its underlying mechanisms. By omitting parameter $\boldsymbol{\theta}$, the Koopman operation $\mathcal{K}$ can be represented as:
\begin{equation}\label{equ4}
	\mathcal{K} g(\boldsymbol{x}_{t}) = g(f (\boldsymbol{x}_{t},\boldsymbol{u}_{t}))=g(\boldsymbol{\dot{x}}_{t}),
\end{equation}
with the DS is represented as:
\begin{equation}\label{equ5}
	\boldsymbol{\dot{x}}_{t}=f (\boldsymbol{x}_{t},\boldsymbol{u}_{t}).
\end{equation}

Utilizing the fitting capabilities of neural networks to learn an approximate Koopman operator typically requires an encoder-decoder architecture. The encoder maps the original data from its native space to a finite-dimensional space where the dynamics can be approximated linearly. Subsequently, the decoder is employed to map the transformed data back to the original space. With a slight abuse of notation, the approximately linear constraint can be enforced by the loss function $||\dot{\varphi}(\boldsymbol{x}_{t})-	\boldsymbol{A}_1\varphi(\boldsymbol{x}_{t})-\boldsymbol{B}_2\boldsymbol{u}_{t}||_2$, where $\varphi(\cdot)$ is the encoder learned during training. Additionally, the decoder, denoted by $\eta (\cdot)$ is employed to output the state through the following loss function $||\boldsymbol{x}_{t}-\eta (\varphi(\boldsymbol{x}_{t}))||_2$. To ensure that the transformation $\eta(\varphi(\boldsymbol{x}_{t}))$
yields the original input 
$\boldsymbol{x}_{t}$, one effective approach is to define 
$\varphi(\boldsymbol{x}_{t})$ as the concatenation of 
$\boldsymbol{x}_{t}$
with a set of learnable features. In this scenario, the function 
$\eta(\cdot)$ performs a truncation operation, which can be effectively represented by multiplication with a constant matrix  $\boldsymbol{C}$, effectively intercepting the input without altering it. Consequently, the decoder of the Koopman neural network does not introduce extra parameters and the loss function $||\boldsymbol{x}_{t}-\eta (\varphi(\boldsymbol{x}_{t}))||_2$ can be omitted, simplifying the training process. In this way, the deep Koopman modeling can be represented as:
\begin{equation}\label{equ6}
	\begin{cases}
		\dot{\varphi}(\boldsymbol{x}_{t})=	\boldsymbol{A}_1\varphi(\boldsymbol{x}_{t})-\boldsymbol{B}_1\boldsymbol{u}_{t} \\
		\boldsymbol{\dot{x}}_{t}=\boldsymbol{C}	\dot{\varphi}(\boldsymbol{x}_{t}).
	\end{cases}
\end{equation}
Which can be rewritten as:
\begin{equation}\label{equ7}
		\boldsymbol{\dot{x}} = \boldsymbol{C}\boldsymbol{A}_1\varphi(\boldsymbol{x})+\boldsymbol{C}\boldsymbol{B}_1 \boldsymbol{u},
\end{equation}
Upon examination of Equation (\ref{equ3}), it is observed that when the function $\varphi(\cdot)$ is defined as a identity operation, as $\varphi(\boldsymbol{x})=\boldsymbol{x}$, they are equivalent.
This equivalence implies that under the specified condition, SSM formulation can be regarded as a particular case of a Koopman operator-based dynamical system. 
Consequently, the Koopman operator exhibits a more robust capacity for expressing non-linear dynamics compared to the traditional SSMs. 
 
However, this study encounters a challenge with the control variable $\boldsymbol{u}$ within the DS remaining unknown, which complicates the evolution of multi-agent states. When using SSM, the nonlinear expressiveness of the proposed algorithm arises solely from the modeling of these control variables $\boldsymbol{u}$. In contrast, when employing the Koopman operator to model the DS, the nonlinear expressiveness derives from both the observable functions $\varphi(\cdot)$ and the control variable $\boldsymbol{u}$. This dual source of nonlinearity may shift the focus of training and impact the representation of potential physical characteristics.

A interesting result is when considering the control variables $\boldsymbol{u}$ as the observable functions $\varphi(\boldsymbol{x})$ in the Koopman operator framework, equation (\ref{equ3}) becomes 
\begin{equation}\label{equ8}
	\boldsymbol{\dot{x}} = (\boldsymbol{x}^{\mathrm{T}} || \varphi^{\mathrm{T}}(\boldsymbol{x}))^{\mathrm{T}}(\boldsymbol{A}^{\mathrm{T}}||\boldsymbol{B}^{\mathrm{T}}),	
\end{equation}
where $||$ denotes concatenate. In this context, the SSM can still be regarded as a specialized case within the Koopman operator framework for predicting multi-agent trajectories. For simplicity, this model is still referred to as an SSM in this paper.

\subsection{Graph mixed Mamba encoder}

Although the dynamics of the multi-agent system are encapsulated within the DS model, the control variables for these agents remain unobserved within the available datasets. Consequently, it is essential to infer the control variables from the historical data using advanced inference techniques. In this context, a neural network encoder is employed to perform this task, leveraging its capability to capture complex patterns and relationships within the data.

The encoder meticulously processes the historical dataset $\mathcal{H}$ , which includes data from all agents, to produce a comprehensive set of representational vectors. These vectors are instrumental for predicting future trajectories. Moreover, these vectors serve as the control variables within the DS model, as discussed in Section \ref{sec4_1}. 
Historical data, which maybe unstructured, can be effectively processed by GNNs. These networks are constructed with multiple layers that operate on all nodes simultaneously. Each layer's operation is node-centric, meaning it focuses on individual nodes within the graph. A key feature of GNNs is the sharing of learnable parameters across all nodes within each layer. This proposed approach incorporates a GAT+ GNN layer \cite{gat3}, which employ an
	attention mechanism to calculate a set of aggregation
	weights across the inclusive neighborhood $\widetilde{N}(v)$ of each node $v$. This approach allows the GNNs to allocate greater importance to specific neighbors within the graph. In this paper, the modified  version is utilized,  the dynamic attention are calculated and the attention weights are determined by:
	\begin{equation}\label{equ9}
	\tilde{\alpha}_{t,v} =\frac{\mathrm{exp}(\boldsymbol{a}^{\mathrm{T}}_{\tilde{\alpha}}\gamma(\boldsymbol{W}_{\tilde{\alpha}}[\boldsymbol{h}^{v}||\boldsymbol{h}^{\tau}||e_{v,\tau}]) )}{\sum_{u \in \widetilde{N}(v) }\mathrm{exp}(\boldsymbol{a}^{\mathrm{T}}_{\tilde{\alpha}}\gamma(\boldsymbol{W}_{\tilde{\alpha}}[\boldsymbol{h}^{v}||\boldsymbol{h}^{u}||e_{v,u}])} ,		
\end{equation}	
where $\boldsymbol{a}_{\tilde{\alpha}}$ and  $\boldsymbol{W}_{\tilde{\alpha}}$   represent learnable parameters that are pivotal for the performance. Function $\gamma(\cdot)$ is utilized to apply the leaky (Rectified Linear Unit) ReLU activation function, which introduces a non-zero gradient for the negative input values. Additionally, $e_{v,u}$ denotes the edge weight that is instrumental in the computation of $\tilde{\alpha}_{t,v}$ and finally, the representation can be derived using the following equation:
	\begin{equation}\label{equ10}
\boldsymbol{h}^{tv} = \boldsymbol{b}+\boldsymbol{W}_{1}\boldsymbol{h}^{v}+\sum_{\tau \in \widetilde{N}(v)}	\tilde{\alpha}_{t,v} \boldsymbol{W}_{2}\boldsymbol{h}^{\tau} ,		
\end{equation}	
GAT+ layers commonly employ multiple attention heads, which are distinct instances of the attention mechanism previously described. Each head operates independently to focus on different facets of the data, thereby generating a set of distinct representation vectors. These vectors capture various aspects of the input data, enhancing the model's ability to learn complex patterns. Subsequently, the individual vectors are combined through concatenation or averaging to form a comprehensive new representation. This process allows the GAT+ layer to integrate information from multiple perspectives, enriching the model's understanding of the input. 
  represents an enhanced iteration of GAT+ architecture. 

 GNNs provide structured feature representations that capture the essence of spatio-temporal dynamics. These features, when conceptualized as time series, offer a robust framework for deciphering the underlying control variables within the DS across temporal dimensions.

 Recently, a novel modeling approach known as Mamba has been introduced for handling sequence information. With a bit abuse of notation, Mamba leverages a linear ODE to model the mapping from an input sequence $x(t) \in \mathbb{R}^{N} $ to an output sequence $y(t) \in \mathbb{R}^{N} $ via a latent state representation $h(t) \in \mathbb{R}^{N} $.
\begin{equation}\label{equ11}
\begin{aligned}
&h'(t) = \boldsymbol{A}_2 h(t)+\boldsymbol{B}_2x(t),	\\
&y(t) = \boldsymbol{C}_2h(t),	
\end{aligned}
\end{equation}
where $\boldsymbol{A}_2 $, $\boldsymbol{B}_2 $, and $\boldsymbol{C}_2 $ represent the state matrix, input matrix, and the output matrix, respectively. While $N$ is a positive integer denoting the dimension of the state.
When operating in continuous time, the output states are often difficult to solve analytically. Meanwhile, the input data are sampled evenly, represent discrete values. The system described by Equation (\ref{equ10}) can be discretized as follows:

\begin{equation}\label{equ12}
	\begin{aligned}
		&\boldsymbol{h}_{t} = \boldsymbol{\bar{A}} \boldsymbol{h}_{t-1}+\boldsymbol{\bar{B}}\boldsymbol{x}_{t},	\\
		&\boldsymbol{y}_{t} = \boldsymbol{C}\boldsymbol{h}_{t},	
	\end{aligned}
\end{equation}
where $\boldsymbol{\bar{A}}=\mathrm{exp}(\bigtriangleup \cdot \boldsymbol{A}_2)$ , $\boldsymbol{\bar{B}}=(\bigtriangleup \cdot \boldsymbol{A}_2)^{-1}(\mathrm{exp}(\bigtriangleup \cdot \boldsymbol{A}_2)-\boldsymbol{I})\cdot \bigtriangleup \cdot \boldsymbol{B}_2$ , $\bigtriangleup$ represents the discretization step size and $\boldsymbol{I}$ is the identity matrix. After the parameters are transformed using a discrete approach, Mamba incorporates a selection mechanism into the model. This mechanism solely affects the interaction along the time series, making $\boldsymbol{A}_2$, $\boldsymbol{B}_2$, and $\boldsymbol{C}_2$ input-dependent. The purpose of this is to compress the time series information into a more compact state representation.

In the context of this paper, the encoder's role is to generate the hidden state vectors, which are assumed to represent the initial control variables. In an effort to enhance the representational capabilities of the Mamba model, this paper introduces a novel mixed Mamba, which integrates diverse features to improve its performance and applicability, the details of which are illustrated in Fig. \ref{fig2}.

The output features produced by the GNNs are represented as $\boldsymbol{x}_{i},$ where 
$i$ ranges from $1$ to $t$. When employing the Mamba block, the calculation of $\boldsymbol{h}_{t}^{v}$ is performed as follows:
\begin{equation}\label{equ13}
\boldsymbol{h}_{t}^{m}=\prod^t_{k=2}\boldsymbol{\bar{A}}_k \boldsymbol{\bar{B}}_1 \boldsymbol{x}_{1} +\prod^t_{k=3}\boldsymbol{\bar{A}}_k \boldsymbol{\bar{B}}_2 \boldsymbol{x}_{2} +...+\boldsymbol{\bar{B}}_t \boldsymbol{x}_{t}
\end{equation}
 Given that $\boldsymbol{A}$ represents the Hippo matrix, which implicitly contain temporal information, and $\boldsymbol{B}$ is a state depend matrix, Mamba can retain all pertinent information. The encoder's primary role is to create a hidden state vector that captures the initial control variables for the DS model, which is essential for accurately forecasting future time series trajectories. To maximize predictive accuracy, it is crucial to fully leverage the temporal historical data. However, the information encoded within a single Mamba block may not be adequate for all applications.
 
 Furthermore, the integration of a mixed Mamba variant is anticipated to significantly enhance the encoder's performance. This innovative approach involves deploying one Mamba block to process the original time series data and a separate Mamba block to handle the inverse time series multiplying with the other time series data. This combined sum is subsequently refined by the final Mamba block, which is expected to bolster the model's predictive accuracy for future time series trajectories.
   
\subsection{Control variable SSM}

In the prediction of future trajectories for multiple agents, their motion patterns are typically not subject to drastic changes that are unpredictable even to human observers. Consequently, the control variables within an approximate physical DS model should also exhibit minimal variation. Such a scenario can be effectively captured using a separate SSM. With a bit abuse of notation, this equation can be expressed as:
\begin{equation}\label{equ14}
	\boldsymbol{u}_{t+1}=\boldsymbol{A}_3\boldsymbol{u}_{t}+\boldsymbol{B}_3g(\boldsymbol{\tilde{x}}_{t},\boldsymbol{u}_{t}),
\end{equation}
Where $\boldsymbol{A}_3$ and $\boldsymbol{B}_3$ are constant matrices learned through  automation, $\boldsymbol{\tilde{x}}_{t}$ denotes the estimated states of the multi-agents derived from the modeled DS. 

\begin{figure}[htbp]
	\centerline{\includegraphics[width=0.5\textwidth]{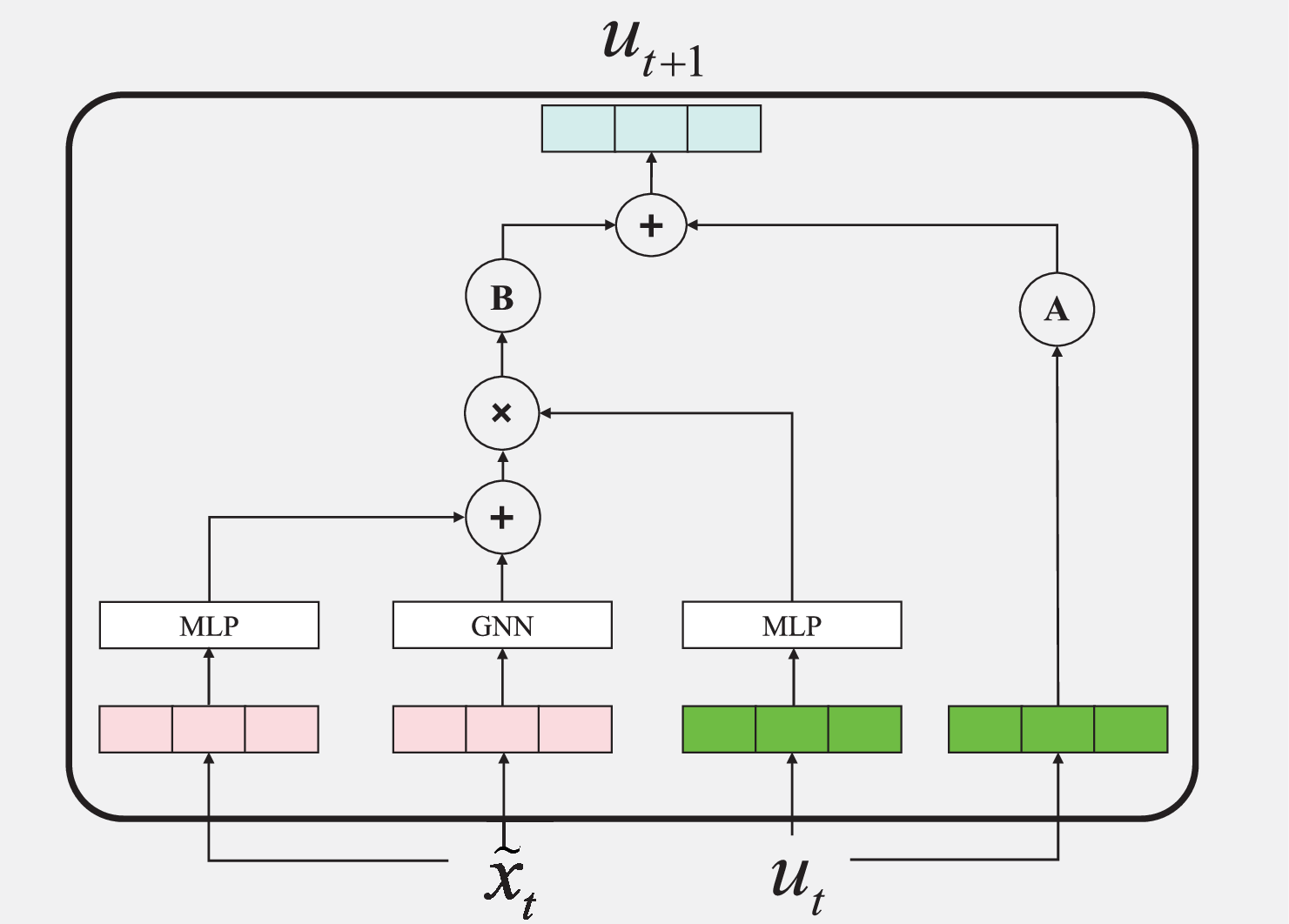}}
	\caption{The designed structure of the evolving control variables.}	
	\label{fig3}
\end{figure}

When  $\boldsymbol{B}_3g(\boldsymbol{x},\boldsymbol{u}_{t})$ is considered as the response to external forces, this model can be viewed as another instance of SSM. The detailed design of the evolution of the control variables from $\boldsymbol{u}_{t}$ and $\boldsymbol{x}$ to  $\boldsymbol{u}_{t+1}$ is illustrated in Fig. \ref{fig3}.

The progressive refinement of control variables enables the model to generate a sequence of features corresponding to different future time steps. This capability enhances the model's ability to forecast future trajectories. Greater accuracy in predicting control variables is directly linked to the precision of trajectory forecasts. Therefore, utilizing more information to refine these control variables can significantly improve the model's predictive performance.

The most commonly used function $g(\boldsymbol{\tilde{x}}_{t},\boldsymbol{u}_{t})$ employs a neural network with $\boldsymbol{\tilde{x}}_{t}$ and $\boldsymbol{u}_{t}$ concatenated as inputs.
While directly incorporating predicted states $\boldsymbol{\tilde{x}}_{t}$
and control inputs 
$\boldsymbol{u}_{t}$
into a neural network does not inherently account for observations or interactions with other agents, accurate trajectory prediction in multi-agent systems requires considering these inter-agent dynamics.

To address this, the graph formulation is reintroduced. GNNs not only evaluate the state of the node representing an agent but also consider the states of neighboring nodes within the graph, capturing spatial relationships and interactions among agents.

By integrating GNNs into the evolution of control variables, the model preserves the spatial properties of the multi-agent system throughout both the encoding and decoding stages. This integration facilitates a more intuitive understanding of the system's dynamics, which is crucial for predicting future trajectories of each agent with greater accuracy.

Therefore, in this paper, $g(\boldsymbol{\tilde{x}}_{t},\boldsymbol{u}_{t})$ integrates estimated states processed by a GNN and a MLP. The outputs from these networks are concatenated and then multiplied by the tensor output from another MLP, which takes the most recent control variables as its input.

However, since future graph features are not directly observable when forecasting future trajectories, the model uses the most recent graph state as a proxy for evolving the control variables. Thus, when predicting future trajectories, the evolution of control variables relies on the current graph information.

The proposed framework for predicting the trajectories of multiple agents can be conceptualized as an encoder-decoder architecture, complemented by a motion model. In this framework, the initial control variables are extracted by the encoder, while the decoder is responsible for their subsequent evolution. The motion model, in turn, employs a SSM to enhance the predictive accuracy.

\subsection{EKF}

Furthermore, the DS within this framework is tasked with estimating process noise. The estimated noise parameters are subsequently integrated into an Extended Kalman Filter (EKF), which serves to augment the precision of trajectory predictions.

This approach is particularly pertinent when dealing with a state-transition function  $f$ 
that is differentiable, along with the consideration of process noise represented by  $\boldsymbol{w}_{k}$
\begin{equation}\label{equ15}
\boldsymbol{x}_{t+1}=\psi(\boldsymbol{x}_{t})+\boldsymbol{w}_{k},
\end{equation}
When using the Euler method and considering the process noise,  the substitution of Equation (\ref{equ8})  into  equation (\ref{equ15})   yields a modified form: 
\begin{equation}\label{equ16}
\boldsymbol{x}_{t+1} =\boldsymbol{x}_{t}+ (\boldsymbol{x}_{t}^{\mathrm{T}} || \varphi^{\mathrm{T}}(\boldsymbol{x}_{t}))^{\mathrm{T}}(\boldsymbol{A}^{\mathrm{T}}||\boldsymbol{B}^{\mathrm{T}})\bigtriangleup  T+\boldsymbol{w}_{k},	\end{equation}
where $\bigtriangleup  T$ denotes a sampling interval and $\psi(\boldsymbol{x}_{t})=\boldsymbol{x}_{t}+ (\boldsymbol{x}_{t}^{\mathrm{T}} || \varphi^{\mathrm{T}}(\boldsymbol{x}_{t}))^{\mathrm{T}}(\boldsymbol{A}^{\mathrm{T}}||\boldsymbol{B}^{\mathrm{T}})\bigtriangleup  T$. To  estimate the process noise using the EKF,
 one must initially compute the Jacobian matrix through a first-order Taylor expansion as: 
\begin{equation}\label{equ17}
	\boldsymbol{F}_{t}=\frac{\partial \psi}{\partial \boldsymbol{x}_t} |_{\boldsymbol{x}_t=\boldsymbol{\hat{x}}_t},
\end{equation}
where $\boldsymbol{\hat{x}_t}$ is the estimated state. The prediction step can be mathematically expressed as follows:
\begin{equation}\label{equ18}
	\begin{aligned}
		&\boldsymbol{\hat{x}_{t+1}}=\psi(\boldsymbol{\hat{x}}_{t}),	\\
		& \boldsymbol{P_{t+1}}=\boldsymbol{F_{t}^{\mathrm{T}}}\boldsymbol{P_{t}}\boldsymbol{F_{t}}+\boldsymbol{Q_{t}}  ,	
	\end{aligned}
\end{equation}
Where $\boldsymbol{Q_{t}}$ represents the process noise, which is typically derived from the decoder's output. The initial state covariance matrix, $\boldsymbol{P_{0}}$, can be initialized as zero matrix. Defining these components enables the EKF to provide an estimation of the model prediction. This capability enhances the performance of the proposed framework by incorporating uncertainty estimates. Utilizing a purely deterministic model for estimating the trajectories of multiple agents typically does not yield optimal performance, as it fails to account for the inherent variability and unpredictability in their movements.

\section{ Evaluation and Results}\label{sec5}
The capabilities of the proposed algorithm have been rigorously assessed through a comprehensive series of evaluations. Initially, the performance of the proposed model was systematically compared against a baseline model across various datasets and scenarios. These comparative analyses, which provide insights into the model's efficacy, are elaborated upon in the subsequent sections. Furthermore, the results of an ablation study are presented, which offer additional clarity on the model's performance by isolating the contributions of individual components.

\subsection{ Evaluation and Results}
To assess the efficacy of the proposed algorithm, a comprehensive evaluation was conducted using three distinct datasets. These datasets, namely highD \cite{b55}, rounD \cite{b56}, were employed throughout both the training and testing phases of the study. The utilized data sets encompass a comprehensive collection of recorded trajectories from diverse locations across Germany. These data sets encompass a diverse range of road types, including highways, intersections, and roundabouts, to ensure a comprehensive representation of real-world driving scenarios. The data were collected over several hours of naturalistic driving, with a sampling frequency of $25 Hz$, which provides a high-resolution record of the driving environment and behavior. The observation window, which dictates the extent of the model's input, is capped at a maximum duration of $3s$.
Following the pre-processing techniques from \cite{b3}, the highD, and rounD data sets have been meticulously pre-processed. Consequently, they consist of $100,404$, $29,248$, and $7,820$ graph sequences, respectively. For the purpose of machine learning model development, the data has been systematically divided: $80\%$ of the total samples are designated for training, $10\%$ for validation, and the remaining $10\%$ allocated for testing.

All implementations were conducted using the PyTorch framework, adhering to the hyperparameter configurations established in the \cite{b3}. Specifically, within the Mamba block, the three key hyperparameters are defined as follows: a state expansion factor of 
$10$ for the SSM , a local convolution width of $4$, and a block expansion factor of $4$. To ensure a fair comparison with prior work, the training process was structured with 
$200$ epochs.
\begin{table}[h]
	\centering
	\tabcolsep=5pt	
	\caption{results}
	\begin{tabular}{lcccccc}
		\hline
		metric & ADE & FDE & MR & APDE & ANLL & FNLL \\
		\hline                       
		highD                            \\
		\hline
		CA  & 0.78 & 2.63 & $0.55$ & 0.73 & \textemdash & \textemdash \\
		CV  & 1.49  & 4.01 & 0.79 &1.89 &\textemdash  &\textemdash  \\
	Seq2Seq & 0.57 & 1.68 & 0.29 & 0.54 & \textemdash & \textemdash \\
	S-LSTM \cite{slstm}  &0.41  &1.49  & 0.22  &0.39 & -0.61 &3.20 \\
		CS-LSTM \cite{cslstm}  & 0.39 & 1.38 &0.19  &0.37  & -0.66 &3.33\\
	GNN-RNN \cite{b41}  & 0.40 & 1.40 & 0.17 & 0.38 & \textemdash & \textemdash \\
		mmTransformer \cite{mmtrans} &0.39  &1.13  &0.15  &0.39 &\textemdash  &\textemdash \\
	Trajectron++ \cite{b9} &0.44 &1.62 & 0.23 &0.32 &-1.57  &1.63 \\
		MTP-GO \cite{b3}  &0.30 &1.07  & 0.13 &0.30 &-1.59  &2.02 \\
	 Proposed algorithm  &$\mathbf{0.26}$ &$\mathbf{0.94}$  & $\mathbf{0.06}$ &$\mathbf{0.25}$ &$\mathbf{-2.12}$  &$\mathbf{1.46}$ \\		
		
		\hline
		roundD                           \\
		\hline
		CA  &4.83 &16.2 & 0.95&3.90  & \textemdash  &\textemdash \\
		CV  & 6.49 & 17.1  &  0.94 &4.34 & \textemdash & \textemdash \\
		Seq2Seq  &1.46  &3.66  & 0.59 &0.82  & \textemdash &\textemdash  \\
		S-LSTM \cite{slstm} &1.20  &3.47  &0.56  &0.74 &1.75  &5.12 \\
		CS-LSTM \cite{cslstm} &1.19  &3.57  &0.60  &0.69 &2.09  &5.54 \\
	GNN-RNN \cite{b41} & 1.29 & 3.50 & 0.59 & 0.77 & \textemdash & \textemdash \\
	mmTransformer \cite{mmtrans} &1.29  & 3.50 &0.59 & 0.77 &\textemdash  &\textemdash \\
	Trajectron++ \cite{b9} &1.09 &3.53  &0.54 &0.59  & $\mathbf{-4.25}$ &$\mathbf{1.50}$\\	MTP-GO \cite{b3} &0.96 &$\mathbf{2.95}$  &0.46  &0.59  & 0.22 &3.38\\
	Proposed algorithm  &$\mathbf{0.93}$ &3.03  & $\mathbf{0.27}$ &$\mathbf{0.57}$ &-0.47  &3.05 \\		
		\hline
	\end{tabular}
	\label{table1}
\end{table}	

\begin{table*}[h]
	\centering
	\tabcolsep=6pt	
	\caption{results}
	\begin{tabular}{lcccc|cccccc}
		\hline
		Dataset & Index & Mixed Mamba & Graph considerd &  Linear Koopman    & ADE & FDE & MR & APDE & ANLL & FNLL \\
		\hline                             
		& $\mathrm{H}_{1}$ & $\checkmark$ & $\times$ & $\checkmark$   & 0.27  &0.97  &0.06 &0.26 &-2.06 &1.51 \\
		
		& $\mathrm{H}_{2}$ & $\times$ & $\checkmark$ & $\checkmark$    &0.27&0.98  &$\mathbf{0.05}$  & 0.26 &-2.02  & 1.49  \\
		\multirow{2}{*}{highD}	 	& $\mathrm{H}_{3}$ & $\checkmark$ &  $\times$ & $\times$    &0.57  &1.27  & 0.15 &0.55  & 0.51 &1.85   \\
		& $\mathrm{H}_{4}$ & $\times$ & $\checkmark$ & $\times$    &$\mathbf{0.24}$  &$\mathbf{0.87}$  &$\mathbf{0.05}$  &$\mathbf{0.23}$  &$\mathbf{-2.47}$  &1.22  \\
		
		& $\mathrm{H}_{5}$ & $\times$ & $\times$ & $\checkmark$    &0.28  &1.00  & 0.06 & 0.26 &-2.03  & 1.53 \\
	& $\mathrm{H}_{6}$ & \checkmark & \checkmark & $\times$    &0.26  &0.91 & $\mathbf{0.05}$  &0.25   &-2.27 &$\mathbf{1.20}$ \\
& $\mathrm{H}_{7}$ & $\times$ & $\times$ & $\times$    &0.27  &1.00  &0.07  &0.26  &-2.41  &1.38 \\

& $\mathrm{H}_{8}$ & \checkmark & \checkmark & \checkmark   &0.25 &0.94  & 0.06 &0.26 &-2.12  &1.46 \\
		\hline                             
		& $\mathrm{H}_{1}$ & $\checkmark$ & $\times$ & $\checkmark$      &1.01  &3.30  &0.30  &0.60 & -0.46 & 3.11 \\
		
		& $\mathrm{H}_{2}$ & $\times$ & $\checkmark$ & $\checkmark$    &  1.05& 3.38 & 0.32 & 0.61 & -0.46 & 3.13  \\
		\multirow{2}{*}{rounD}	 	& $\mathrm{H}_{3}$ & $\checkmark$ & $\times$ & $\times$    & 1.01 &3.29 &0.32  & 0.59 & -0.64 &  3.14 \\
		& $\mathrm{H}_{4}$ & $\times$ & $\checkmark$ & $\times$   &1.00  &3.24  & 0.33 &0.60  & $\mathbf{-0.74}$ &$\mathbf{2.94}$  \\
		
		& $\mathrm{H}_{5}$ & $\times$ & $\times$ & $\checkmark$   &1.05  &3.40 &0.32 & 0.61 & -0.40 & 3.20 \\
		& $\mathrm{H}_{6}$ & \checkmark & \checkmark & $\times$    &0.97  &3.19 & 0.29  & $\mathbf{0.56}$  &-0.65 & 3.03\\
			& $\mathrm{H}_{7}$ & $\times$ & $\times$ & $\times$    &1.05  &3.39  &0.34  &0.62  & -0.64 & 2.99\\
			
			& $\mathrm{H}_{8}$ & \checkmark & \checkmark & \checkmark   &$\mathbf{0.93}$  &$\mathbf{3.03}$  &$\mathbf{0.27}$  &0.57  &-0.47  &3.05 \\
		\hline             
	\end{tabular}
	\label{table2}
\end{table*}
 To thoroughly assess the predictive capabilities of the proposed algorithm, a carefully curated set of six distinct metrics was employed. These metrics were meticulously chosen to provide a robust evaluation framework. 
\begin{itemize}
	\item Average Displacement Error (ADE): The average $L_2$-norms between the predicted and actual trajectories over the entire prediction horizon. Mathematically, ADE is computed as follows:
	\begin{equation}\label{equ18}
		\mathrm{ADE}=\frac{1}{t_f}\sum_k=1^t_f ||\hat{\boldsymbol{x}}_k-\boldsymbol{x}_k||_2.
	\end{equation}
	where $t_f$ is the total prediction horizon, $\hat{\boldsymbol{x}}_k$ is the estimated state from the model while $\boldsymbol{x}_k$ denotes the corresponding true state from the dataset. This metric assesses the mean accuracy of predictions.
	
	\item  Final Displacement Error (FDE): FDE is defined by the $L_2$-norm of the discrepancy between the final predicted position and the actual position. 
	\begin{equation}\label{equ19}
		\mathrm{FDE}= ||\hat{\boldsymbol{x}}_{t_f}-\boldsymbol{x}_{t_f}||_2.
	\end{equation}
	This metric serves as a measure of the model's predictive accuracy concerning events situated at a considerable temporal remove.
	\item Miss Rate (MR): This metric represents the proportion of instances in which the predicted final position deviates significantly from the ground truth, which is defined as a threshold of $2m$. The MR is indicative of the consistency of the predictive model's performance.
	\item The Average Path Displacement Error (APDE): This metric is calculated as the mean between the predicted positions and the minimum $L_2$-norms corresponding ground truth value.
	\begin{equation}\label{equ20}
		\begin{aligned}
			&\mathrm{APDE}=\frac{1}{t_f}\sum_k=1^t_f ||\hat{\boldsymbol{x}}_k-\boldsymbol{x}_{k^*}||_2,	\\
			& k^* = \underset{i}{\mathrm{argmin}}||\hat{\boldsymbol{x}}_k-\boldsymbol{x}_i||_2,	
		\end{aligned}
	\end{equation}
This metric is utilized to quantify the error associated with the trajectory predictions and serves as an index of the accuracy of the predicted maneuvers.
	\item  Average Negative Log-Likelihood (ANLL): ANLL is a statistical measure that quantifies the goodness of fit between a predicted probability distribution and the observed data. It is calculated as the mean of the negative log-likelihoods across all observations, offering an aggregate assessment of the model's predictive performance.
		\begin{equation}\label{equ21}
		\mathrm{ANLL}= -\frac{1}{t_f} \sum_{k=1}^{t_f} \mathrm{log}(\mathcal{N}(\boldsymbol{x}_k|\boldsymbol{x}_k,\boldsymbol{P}_k)) .
	\end{equation}
	\item Final Negative Log-Likelihood (FNLL): This metric represents Negative Log-Likelihood evaluated at the final step of the total prediction horizon. which has the similar metric performance as FDE.
		\begin{equation}\label{equ22}
		\mathrm{FNLL}= - \mathrm{log}(\mathcal{N}(\boldsymbol{x}_{t_f}|\boldsymbol{x}_{t_f},\boldsymbol{P}_{t_f})) .
	\end{equation}
\end{itemize}

The following models are implemented for performance comparison:
\begin{itemize}
	\item Constant Acceleration (CA):
An open-loop model that assumes constant acceleration.
	\item Constant Velocity (CV):
An open-loop model that assumes constant velocity.
	\item Sequence to  Sequence (Seq2Seq):  A baseline LSTM-based encoder-decoder model.
\item Social LSTM (S-LSTM) \cite{slstm}: An encoder-decoder network based on LSTM for trajectory prediction, where interactions are encoded using social pooling tensors.
	\item Convolutional Social Pooling (CS-LSTM) \cite{cslstm}: 	Similar to S-LSTM, but learns interactions using a convolutional neural network (CNN).
	\item Graph Recurrent Network (GNN-RNN) \cite{b41}:	Encodes interactions using a graph network and generates trajectories with an RNN-based encoder-decoder.
	\item mmTransformer \cite{mmtrans}:
	A transformer-based model for multimodal trajectory prediction, with interactions encoded using multiple stacked transformers.
	\item Trajectron++ \cite{b9}: A GNN-based recurrent model that performs trajectory prediction using a generative model combined with hard-coded kinematic constraints.
		\item MTP-GO \cite{b3}:
	Utilizes GNN-GRU for both the encoder and decoder to extract features, which are then used by a neural ODE to predict trajectories.
\end{itemize}

The S-LSTM, CS-LSTM, MTP-GO, and the proposed algorithm uniquely allow for the analytical computation of the NLL, a feature not available with the sampling-based Trajectron++. Although the NLL values from Trajectron++ may not perfectly reflect the true likelihood, they are obtained using a kernel density estimate derived from samples of the predictive distribution. Additionally, non-probabilistic metrics for Trajectron++ are averaged over samples from the most probable component. Similarly, metrics for the mmTransformer are calculated based on the trajectory with the highest confidence score.

The comparative results are presented in Table \ref{table1}, which utilizes the highD and rounD datasets. The metrics reported represent the average outcomes for all vehicles across various traffic scenarios. An analysis of Table \ref{table1} reveals that the proposed algorithm has achieved nearly optimal results across all datasets when evaluated using a comprehensive set of metrics.

The proposed model exhibits its optimal performance on the highD dataset and nearly optimal performance in comparison on the rounD dataset. This disparity in performance maybe be attributed to the nature of the prediction tasks. Highway trajectory prediction is arguably less complex due to the predominantly linear trajectories observed among vehicles on highways. Additionally, the highD dataset benefits from a larger volume of data, which likely contributes to the model's enhanced performance. Conversely, roundabout trajectory prediction poses a greater challenge. This complexity arises from the continuous variations in the velocity and acceleration of the agents involved, which introduces greater challenge into the prediction process. The vehicle motion patterns in the highD dataset are relatively straightforward compared to those in the rounD dataset. The control variables in the highD dataset are more easily estimated using the proposed algorithm. Consequently, the proposed method is expected to achieve superior performance.
In the rounD dataset, the proposed algorithm does not achieve optimal performance in terms of the FDE metric. This may be attributed to the estimation of control variables using a uniform model pattern, which could introduce errors at each estimation step. Consequently, the cumulative effect of these errors may result in a significant final displacement error. 

The proposed method comprises three main components: a SSM to represent the motions of multiple agents, a mixed Mamba algorithm to estimate the initial control variables, and a decoder to further evolve these control variables. To gain a deeper understanding of the impact of these components on predictive performance, an ablation study was conducted. The primary aim of this study was to systematically evaluate the model's architecture, with the goal of pinpointing the key mechanisms that most significantly contribute to its overall performance

The ablation study encompassed two distinct datasets: highD and rounD.  The study meticulously examined three integral components to ascertain their individual and collective influence on the model's efficacy.
\begin{itemize}
	\item Comparing the use of a single Mamba with the use of a mixed Mamba: This paper proposes the utilization of a mixed Mamba structure to obtain the initial control variables. To assess the efficacy of this mixed Mamba structure, a comparative analysis was conducted, where all other components of the system remained, with the sole exception of the transition from a mixed Mamba to a single Mamba structure.
	
	\item This study evaluates the evolution of control variables when graph information is incorporated, in contrast to scenarios where it is disregarded: This study estimates control variables for trajectory prediction by incorporating the graph information from the last state into all subsequent control estimates. To evaluate the effectiveness of incorporating graph information, a comparative analysis was conducted. In this analysis, all other system components were kept, while the control variable estimation process was modified to exclude the use of graph information.

	\item The study contrasts the complexities of non-linear state modeling against the structured simplicity of DS control's linear state modeling: In the preceding section, this paper introduced the use of equation \ref{equ3} to represent the DS for multi-agent systems, aiming to encapsulate all non-linear dynamics within the control variables. In contrast, the Koopman operator-based model, as defined by equation \ref{equ7}, also captures non-linearity in the state representation. To assess the impact of incorporating non-linearity solely in the control variables, a comparative analysis was performed (denoted as linear Koopman). This analysis maintained all other system components constant, while the DS modeling was adapted to utilize equation \ref{equ7} (denoted as non-linear Koopman).
\end{itemize}

As depicted in Table \ref{table2}, the three components have distinct impacts on the two datasets. It is evident that incorporating graph information enhances the model's performance across both datasets. However, the influence of the other components varies, suggesting that their contributions are dataset-specific.

 Table \ref{table2} reveals that the proposed structure did not uniformly excel across all metrics when applied to the highD dataset. Specifically, the $H_4$ configuration, which utilizes a single Mamba for the preliminary extraction of control variables and an non-linear Koopman model, nearly achieved the best performance. Directly assessing the impact of Mamba and Koopman components is complex due to their coupling, non-linear interactions that contribute to performance, as illustrated in Table \ref{table2} for the highD dataset.
 
 Several hypotheses may help explain these findings. When employing multi-agent modeling within the SSM framework, denoted as a linear Koopman model, the dynamic models do not fully account for the diverse accelerations of individual agents. A comparative analysis of the CA and CV models, as detailed in Table \ref{table1}, shows that the CA model significantly outperforms the CV model when applied to the highD dataset. This suggests that considering acceleration may be crucial for enhancing model performance. 
 
 The non-linear Koopman model may implicitly account for acceleration, which could contribute to its performance. The introduction of high non-linearity into the model could be a key factor. Meanwhile, the simpler Mamba component may reduce model complexity, thereby mitigating the risk of overfitting and potentially enhancing performance. The $H_3$ configuration, which exhibits the poorest performance, might be suffering from overfitting issues.
 
 In contrast, configurations $H_3$ and $H_6$, which incorporate graph information, achieve significantly better performance. This suggests that the graph information might serve a latent regularization function, helping to prevent overfitting. 
 
 Your paragraph provides a detailed analysis of the results obtained from the rounD dataset. However, it can be made more coherent and readable with some adjustments. Here's a revised version:
 
 "When analyzing the rounD dataset, the CA and CV models yield comparable results, suggesting that a linear Koopman model may suffice for predicting multi-agent trajectories in this context. Table \ref{table2} indicates that the inclusion of both the mixed Mamba model and graph information enhances the model's performance. However, the impact of the linear Koopman model varies.
 
 The non-linear Koopman model, in particular, significantly improves performance across both the ANLL and FNLL metrics. When considering the three components individually, the linear Koopman model does not exhibit a markedly positive effect. Yet, when combined with the mixed Mamba model, it achieves the best performance in terms of ADE, FDE, and MR.
 
 This enhanced performance may be attributed to the fact that the SSM does not inherently incorporate non-linear expressions. Therefore, a complex encoder, as provided by the mixed Mamba model, could be instrumental in bolstering the model's expressive capacity. 
 
 Furthermore, the interactions among agents in the rounD dataset are more complex than those in the highD dataset. In simpler scenarios, such as the highD dataset, the use of the mixed Mamba model might lead to overfitting compared to using a single Mamba model. In contrast, in more complex states, as seen in the rounD dataset, the mixed Mamba model could have a positive impact on the model's performance.
 
 This suggests that the mixed Mamba model's contribution to the model's performance is context-dependent. In less complex environments, its additional complexity may not be necessary and could potentially lead to overfitting. However, in more intricate interaction settings, the mixed Mamba model's ability to capture nuanced dynamics could enhance the predictive accuracy of the model.

\section{Conclusions and Future Work}\label{sec6}
In this paper, a novel framework is introduced for predicting multi-agent trajectories. The framework is developed in three stages: First, a SSM is constructed to represent the dynamics of multiple agents. Second, a mixed Mamba model is proposed to extract initial control variables for the SSM. Finally, a control variable evolution model is formulated to infer the control variables at each step. The efficacy of this framework is rigorously tested across various datasets, demonstrating its superiority over existing state-of-the-art methods in multiple performance metrics. These results highlight the framework's applicability and effectiveness in real-world traffic scenarios.

However, the current SSM does not account for the effects of acceleration or other physical dynamics. Future work will explore models that incorporate these factors, potentially enhancing the framework's predictive accuracy and robustness.

\end{document}